# Learning Depth from Monocular Videos using Direct Methods

Chaoyang Wang, José Miguel Buenaposada, Rui Zhu, Simon Lucey
Carnegie Mellon University

## Abstract

*The ability to predict depth from a single image - using recent advances in CNNs - is of increasing interest to the vision community. Unsupervised strategies to learning are particularly appealing as they can utilize much larger and varied monocular video datasets during learning without the need for ground truth depth or stereo. In previous works, separate pose and depth CNN predictors had to be determined such that their joint outputs minimized the photometric error. Inspired by recent advances in direct visual odometry (DVO), we argue that the depth CNN predictor can be learned without a pose CNN predictor. Further, we demonstrate empirically that incorporation of a differentiable implementation of DVO, along with a novel depth normalization strategy - substantially improves performance over state of the art that use monocular videos for training.*

## 1. Introduction

Depth prediction from a single image using CNNs has had a surge of interest in recent years [7, 21, 20]. Recently, unsupervised methods that rely solely on monocular video for training (without depth or stereo groundtruth) have captured the attention of the community. Of particular note in this regard is the work of Zhou et al. [31] who proposed a strategy that learned separate pose and depth CNN predictors by minimizing the photometric consistency across monocular video datasets during training. Although achieving impressive results this strategy falls noticeably behind those that have been trained using rectified stereo image pairs [13, 19]. These rectified stereo methods have shown comparable accuracy to supervised methods [26, 7, 23] over datasets where only sparse depth annotation is available. However, the assumption of using calibrated binocular image pairs excludes itself from utilizing monocular video which is easier to obtain and richer in variability. This performance gap between stereo [13, 19] and monocular [31] learning strategies is of central focus in this paper.

In this paper, we attempt to close this gap by drawing inspiration from recent advances in direct visual odome-try. Specifically, we note that the major difference between stereo and monocular strategies stems from: (i) unknown camera pose between frames, and (ii) ambiguity in scale. Existing methods [28, 31] for learning depth from monocular video address these differences only partially by adding an extra CNN pose prediction module. In this paper we argue that these previous strategies do not adequately address the scale ambiguity issue - causing divergence during training. Further, we advocate that the incorporation of an additional CNN pose prediction module is unnecessary. Instead we argue that one could employ a differentiable and deterministic objective for pose prediction which is now commonly employed within the SLAM community for direct visual odometry [10, 8, 6, 2].

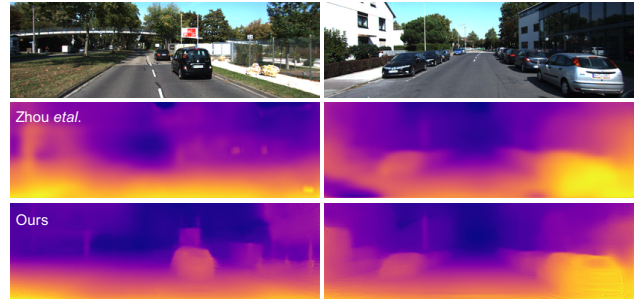

Figure 1. Our depth prediction (3rd row) compared against Zhou *et al.* [31] (2nd row) on KITTI dataset. Our method recovers more fine details such as tree trunks and advertising board.

**Contributions:** We make the following contributions. First, we characterize theoretically and demonstrate empirically why scale ambiguity in current monocular methods is problematic. Specifically, the problem lies in the scale sensitivity of the depth regularization terms employed during training. Inspired by related work in direct visual odometry [10] we propose a simple normalization strategy that circumvents many of these problems and leads to noticeably improved performance over Zhou *et al.* [31](see Fig. 1 for qualitative comparison).

Second, we suggest that learning an additional pose predicting CNN (which we shall refer to herein as the Pose-CNN) is not the most effective strategy for estimating a depth predicting CNN from monocular video. The Pose-CNN employed by Zhou *et al.* does not fully exploit the relation between camera pose and depth prediction, and ignores the fact that pose estimation from depth is a well-studied problem with known geometric properties and well performing algorithms. We instead propose the incorporation of a Direct Visual Odometry (DVO) [27] pose predictor into our framework as: (i) it requires no learnable parameters, (ii) establishes a direct relationship between the input dense depth map and the output pose prediction, and (iii) it is derived from the same image reconstruction loss used for minimizing our entire unsupervised training pipeline. To incorporate DVO into end-to-end training, we propose a differentiable implementation of the DVO (DDVO module), so that the back-propagation signals reaching the camera pose predictor can be propagated to the depth estimator.

Finally, since DVO is a second order gradient descent based method, a good initialization point can lead to better solution. So instead of starting our DDVO module from the identity pose, we propose a mixed training procedure - use a pretrained Pose-CNN to provide pose initialization for DDVO during training. We demonstrate empirically that this hybrid method provides better performance compared to training with Pose-CNN or DDVO alone, and achieves comparable results to Gordard *et al.* [13], which is the state of the art method trained with calibrated binocular pairs on KITTI dataset [12].

**Notation:** lowercase boldface symbols (*e.g.* $\mathbf{x}$) denote vectors, uppercase boldface symbols (*e.g.* $\mathbf{W}$) denote matrices, and uppercase calligraphic symbols (*e.g.* $\mathcal{I}$) denote images. We also use the notations $\mathcal{I}(\mathbf{x}) : \mathbb{R}^2 \rightarrow \mathbb{R}^K$ to indicate sampling of the $K$-channel image representation at subpixel location $\mathbf{x} = [x, y]^\top$.

## 2. Learning depth estimation from videos

Our goal is to learn a function $f_d$ (modeled as a CNN)- parametrized by $\theta_d$- which predicts the inverse depth map $\mathcal{D}$ from a single image $\mathcal{I}$. Instead of doing supervised learning, we want to learn from more widely available data sources without groundtruth depth. Moreover, instead of restricting ourselves to calibrated binocular images [11, 13, 19], we go for more general case of monocular video sequences.

Before introducing our end-to-end training pipeline, it is worth to mention an alternative approach. Given the temporal cues between sequential video frames, it is possible to first get auxiliary depth annotation from some structure-from-motion (SfM) algorithm, and then use its output as supervision to learn the depth estimator. Suppose that the SfM algorithm can be oversimplified as doing photometric bundle adjustment, which minimizes a cost function combining appearance error $\mathcal{L}_{\text{ap.}}$, measuring dissimilarity of pixel-wise correspondences, and some prior cost $\mathcal{L}_{\text{prior}}$, encouraging the smoothness of the depth map. This procedure can be summarized as a two-step optimization:

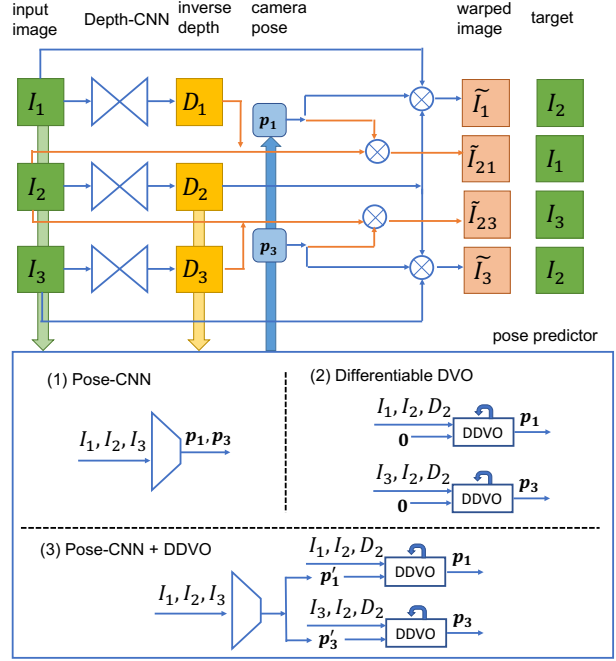

Figure 2. Illustration of our unsupervised learning pipeline. The learning algorithm takes three sequential images at a time. The Depth-CNN produces three inverse depth maps for the inputs, and the pose predictor (lower part) estimates two relative camera pose between the second image and the other two. The appearance dissimilarity loss is measured between the second image $\mathcal{I}_2$ and the inversely warped images of $\mathcal{I}_1, \mathcal{I}_3$; In addition, the loss is evaluated in a reverse direction (marked by orange arrows) - it is also measured between $\mathcal{I}_1, \mathcal{I}_3$ and two warped images of $\mathcal{I}_2$. Lower part of the figure illustrates three architectures we evaluated for pose prediction: 1) Pose-CNN, 2) use our proposed differentiable Direct Visual Odometry (DDVO), the initialization of pose is set as zero (identity transformation), and 3) a hybrid of the above two - use pretrained Pose-CNN to give a better initial pose for DDVO.

1. Compute inverse depth maps and camera poses with SfM:
$$\mathcal{D}^*, \mathbf{p}^* = \arg\min_{\mathcal{D},\mathbf{p}} \mathcal{L}_{\text{ap.}}(\mathcal{D}, \mathbf{p}) + \mathcal{L}_{\text{prior}}(\mathcal{D}) \quad (1)$$

2. Use inverse depth maps from Eq. 1 as supervision to learn the depth estimator:
$$\min_{\theta_d} \mathcal{L}(f_d(\mathcal{I}; \theta_d), \mathcal{D}^*) \quad (2)$$

We argue that this two-step optimization is sub-optimal in theory. Since we are assuming that the SfM's cost function in Eq. 1 reflects the quality of the depth map prediction, what we really want is to find $\theta_d$ which leads to the minimum of this cost. However, minimizing Eq. 2 in the second step does not necessarily lead to minimizing Eq. 1.

Therefore, end-to-end learning of a depth predictor by directly minimizing the cost function in Eq. 1 is, in princi-ple, the best approach. In practise, however, since Eq. 1 is an oversimplification of a real SfM pipeline, it is still possible to get better supervision from the above two-step optimization approach compared to an end-to-end training with gradient descent. Hence, as covered in section 3, one of our contribution is to bring one piece of the sophistication of modern SLAM algorithms - *e.g.* direct visual odometry into the training framework.

Overall, the end-to-end training objective is written as:

$$\min_{\theta_d, \mathbf{p}} \mathcal{L}_{\text{ap.}}\left(f_d\left(\mathcal{I}; \theta_d\right), \mathbf{p}\right) + \mathcal{L}_{\text{prior}}\left(f_d\left(\mathcal{I}; \theta_d\right)\right) \quad (3)$$

Taking apart the joint minimization, we have the following equivalent formulation:

$$\min_{\theta_d} \min_{\mathbf{p}} \mathcal{L}_{\text{ap.}}\left(f_d\left(\mathcal{I}; \theta_d\right), \mathbf{p}\right) + \mathcal{L}_{\text{prior}}\left(f_d\left(\mathcal{I}; \theta_d\right)\right) \quad (4)$$

Minimizing $\mathcal{L}_{\text{ap.}}$ over camera pose $\mathbf{p}$ in Eq. 4 can be viewed as a function which takes inputs from sequential images $\mathcal{I}_1, \ldots, \mathcal{I}_n$ and current depth estimation:

$$f_p(\mathcal{D}, \mathcal{I}_1, \ldots, \mathcal{I}_n) \triangleq \arg\min_{\mathbf{p}} \mathcal{L}_{\text{ap.}}\left(\mathcal{D}, \mathbf{p}\right). \quad (5)$$

Here, we omit $\mathcal{I}_i$ from the input of $\mathcal{L}_{\text{ap.}}$ for conciseness, and we refer $f_p$ as an auxiliary pose predictor. Substitute Eq. 5 to the end-to-end training objective function(Eq. 4), we have our final formulation of the training objective:

$$\min_{\theta_d} \mathcal{L}_{\text{ap.}}\left(f_d\left(\mathcal{I}; \theta_d\right), f_p\left(f_d(\mathcal{I}; \theta_d)\right)\right) + \mathcal{L}_{\text{prior}}\left(f_d\left(\mathcal{I}; \theta_d\right)\right). \quad (6)$$

Detailed definition of our appearance and prior smoothness loss is given in section 4.

Optimizing this loss function meets two crucial issues, which we will address over the following subsections.

### 2.1. Scale ambiguity

Unlike training from calibrated binocular pairs with known baseline, estimating depth and pose from monocular sequence bears scale ambiguity. Any two inverse depth maps differing only in scale (with the pose scaled correspondingly) are equivalent in the projective space. Hence, they have the same appearance dissimilarity loss, or in other words, $\mathcal{L}_{\text{ap.}}$ is scale-invariant in this sense. However, for the regularization term $\mathcal{L}_{\text{prior}}$, usually formulated with derivatives of the inverse depth map [19, 31], smaller inverse depth scale always results in smaller loss value. This scale-sensitive property of the regularization term leads to a catastrophic effect, the loss function in Eq. 6 does not even have local minima, let alone global minima. The proof for this is simple: given any inverse depth estimation, we can construct a new inverse depth map through decreasing its scale (and updating the pose accordingly), which always results in lower loss.

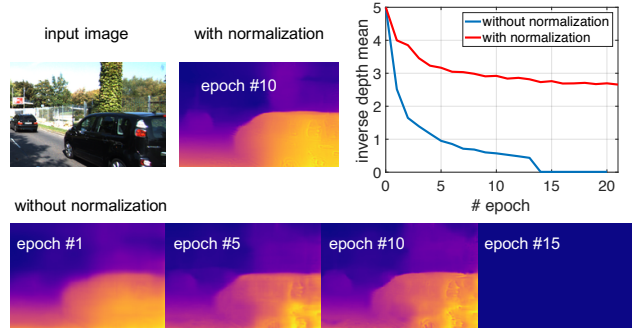

Figure 3. Effect of inverse depth map normalization. top right: Without normalization, the scale (blue curve, quantified by the mean of inverse depth) decreases throughout each epochs during training, and finally saturates to zero; with normalization, the scale stabilizes within a reasonable range(red curve); bottom: Due to the decrease in scale, the prior loss penalizes less on the smoothness of the depth estimation. As a result, the model makes increasingly more structure mistakes. And the training finally breaks before the 15th epoch, where the prediction degenerates to zero; top mid: With normalization, the depth prediction is more precise.

Surprisingly, none of the previous works [28, 31] address this issue. Empirically, we find that this critical defect of the loss function not only leads to inferior results compared to training with binocular pairs, but also results in divergence during training. As shown in Fig. 3, with our re-implementation of Zhou *et al.*'s architecture, the scale of inverse depth prediction is decreasing throughout each training epoch. After several epochs, the training diverges and finally breaks due to the inverse depth prediction saturates to close to zero.

To solve this problem, we propose a simple yet effective approach - apply a non-linear operator to normalize the output of the depth CNN before feeding it to the loss layer. The operator $\eta(\cdot)$ we use here is dividing the input tensor by its mean:

$$\eta(d_i) \triangleq \frac{N d_i}{\sum_{j=1}^{N} d_j} \quad (7)$$

Although this normalization trick has also been used in LSD-SLAM [10] for keyframe selection, it has not been used in the context of learning depth estimation from monocular videos. Empirically we find that this simple normalization trick significantly improves results (see Fig. 3) as it removes the scale sensitive problem of the loss function.

### 2.2. Modeling pose predictor

In previous works, the pose predictor is approximated as a feed-forward CNN, which takes input only from a sequence of frames. Though this approach enjoys the simplicity of using a black box, and is easy to integrate into end-to-end training, it ignores the geometric relation between camera pose and depth as summarized in Eq. 5 which could be exploited for better training.

We propose an alternative approach which attempts to close this discrepancy between theory and practice through noticing the fact that Eq. 5 relates to a class of well-studied geometry-based methods, *e.g.* direct visual odometry (DVO). Given the current depth estimation from $f_d$, DVO solves for camera pose through minimizing the photometric reprojection error between a reference image and a source image. DVO relates to a more general class of image registration method – the Lucas-Kanade(LK) algorithm. It applies the same Gauss-Newton minimization as in LK for faster convergence, and use the inverse composition trick for computational efficiency.

However, to incorporate DVO as the pose prediction module $f_p$ in the end-to-end training, we need to be able to evaluate the derivatives of DVO. Inspired by recent differentiable implementation of inverse compositional LK for object tracking [29] and planar alignment [4], we propose a (sub)differentiable DVO implementation which is described in detail in section 3. With it, during training, the gradient of estimated depth with respect to loss $\mathcal{L}_{\text{ap.}}$ comes from two sources: partial derivative of loss over depth and partial derivative of loss over pose:

$$\frac{d\mathcal{L}_{\text{ap.}}}{df_d} = \frac{\partial \mathcal{L}_{\text{ap.}}}{\partial f_d} + \frac{\partial \mathcal{L}_{\text{ap.}}}{\partial f_p} \frac{\partial f_p}{\partial f_d}. \quad (8)$$

It is worth to mention that Eq. 8 demonstrates one of our major theoretical difference to Zhou *et al.* [31] - our depth estimator gets additional back-propagation signals from the pose prediction, while they ignore the role depth estimation plays in getting camera pose.

We experiment with two versions of applying our differentiable direct visual odometry (DDVO) module in training: 1) we train our Depth-CNN from scratch with DDVO; 2) we first pretrain the Depth-CNN with Pose-CNN, then we fine-tune the Depth-CNN by using the output of Pose-CNN as initialization for DDVO. The motivation for the second approach can be explained in both ways: compared to initializing from the identity pose, Pose-CNN provides a better starting point for DVO; on the other hand, DVO refines the output of Pose-CNN with geometric cues.

## 3. Differentiable direct visual odometry

### 3.1. Direct visual odometry

Steinbrücker *et al.* [27] first proposed the direct visual odometry in the form of minimizing pixel-wise photometric error between sequential RGB-D frames. It was later improved with more robust cost functions [15, 17] and features [1], and extended to photometric bundle adjustment [6, 2, 10]. Compared to visual odometry methods using sparse feature correspondences, direct methods offer better accuracy when inter-frame motion is small, since they make use of the whole dense information contained in the image. Hence direct methods have been widely used in current state-of-the-art visual SLAM systems [10, 9].

Direct visual odometry takes input from a reference image $\mathcal{I}$, its corresponding depth map $\mathcal{D}$, and a source image $\mathcal{I}'$. Let the intensity and inverse depth of a pixel coordinate $\mathbf{x}_i \in \mathbb{R}^2$ at the reference image be respectively stated as $\mathcal{I}(\mathbf{x}_i)$ and $d_i$; camera pose be represented by the concatenation of translation $\mathbf{t} \in \mathbb{R}^3$ and exponential coordinates $\omega \in \mathbb{R}^3$; and the camera intrinsics be known, the projection from the reference image coordinates $\mathbf{x}_i$ to source image coordinates $\mathbf{x}'_i$ is computed as:

$$\mathbf{x}'_i = \mathcal{W}(\mathbf{x}_i; \mathbf{p}, d_i) \triangleq \langle \mathbf{R}\tilde{\mathbf{x}}_i + d_i \mathbf{t}\rangle, \quad (9)$$

where rotation matrix $\mathbf{R}$ is computed from exponential coordinates $\omega$ by the *Rodrigues' formula* [25]; $\tilde{\mathbf{x}}_i$ is the homogeneous coordinate for $\mathbf{x}_i$; and $\langle \cdot \rangle$ projects 3D points to the image plane:

$$\left\langle [x\ y\ z]^T \right\rangle = \left[\frac{x}{z}\ \frac{y}{z}\right]^T \quad (10)$$

The objective of direct visual odometry is to find an optimum camera pose $\mathbf{p}$ which minimizes the photometric error between the warped source image and the reference image which is assumed to be always at the identity pose $\mathbf{0}$:

$$\min_{\mathbf{p}} \sum_i ||\mathcal{I}'(\mathcal{W}(\mathbf{x}_i; \mathbf{p}, d_i)) - \mathcal{I}(\mathcal{W}(\mathbf{x}_i; \mathbf{0}, d_i))||^2. \quad (11)$$

This nonlinear least square problem can be solved efficiently through Gauss-Newton method. For computational efficiency, instead of linearly approximating the source image as in vanilla Gauss-Newton method, the Inverse Compositional algorithm [3] reverses the roll of source image and reference image and compute the parameters update for the reference image. As a result, the Jacobian and Hessian matrix do not need to be re-evaluated per iteration. We summarize the algorithm as following:

1. Precompute the Jacobian matrix, and its pseudo inverse:

$$\mathbf{J} = \begin{bmatrix} \vdots \\ \nabla \mathcal{I}(\mathbf{x}_i) \cdot \left.\frac{\partial \mathcal{W}(\mathbf{x}_i; \mathbf{p}, d_i)}{\partial \mathbf{p}}\right|_{\mathbf{0}} \\ \vdots \end{bmatrix}, \quad \mathbf{J}^\dagger = \left(\mathbf{J}^T \mathbf{J}\right)^{-1} \mathbf{J}^T. \quad (12)$$

2. Warp the source image by the current estimate of camera pose $\mathbf{p}$, and convert it to a vector denoted as $\mathbf{I}'_{\mathbf{p}}$. Form a binary diagonal matrix $\mathbf{W}$ whose $i_{\text{th}}$ diagonal element represents if projected coordinate $\mathbf{x}'_i$ is in view. We use this weight matrix to exclude out-of-view pixels from taking part in computing pose update.

3. Compute pose update:

$$\Delta \mathbf{p} = \mathbf{J}^\dagger \mathbf{W}(\mathbf{I} - \mathbf{I}'_\mathbf{p}), \tag{13}$$

and compose it to get a new pose estimation:

$$\mathbf{p} \leftarrow \Delta \mathbf{p} \circ \mathbf{p}, \tag{14}$$

here, we define the operator ∘ such that it applies a left multiplicative update to the pose.

4. Return to step 2, and repeat till converge.

The main restriction of direct methods is that inter-frame motion must be small. To handle larger motions, we form pyramids of images and depths map through downsampling, and update the pose estimate through coarse-to-fine approach.

### 3.2. Differientiable implementation

Our differientiable implementation of DVO is similar to the inverse compositional spatial transformer network [22]. Both methods iteratively update the geometric transformation parameters and inversely warp the source image through bilinear sampling. The difference is that, instead of using regression network layer with learnable parameters, our "regressor" is deterministically formed with Eq. 13. Computing derivatives with respect to Eq. 13 involves differentiating matrix pseudo-inverse, which we follow similar implementation as in [29].

## 4. Training loss

Our training loss is a combination of appearance loss and prior smoothness loss, aggregated through 4 scales. The appearance loss measures dissimilarity of pixel-wise correspondence between a triplet of images(illustrated in Fig. 2); and the prior loss enforce smoothness for the inverse depths $\mathcal{D}_i$ of the images in the triplet. The loss on the $k$th scale is expressed as:

$$\mathcal{L}^{(k)} = \sum_{i \neq j \in \{1,2,3\}} \mathcal{L}_{\text{ap.}}(\mathcal{D}_i^{(k)}, \mathbf{p}_{ij}; \mathcal{I}_i^{(k)}, \mathcal{I}_j^{(k)}) \\ + \lambda \sum_{i=1}^{3} \mathcal{L}_{\text{prior}}(\mathcal{D}_i^{(k)}). \tag{15}$$

We use weighting parameter $\lambda$ to control the penalty from smoothness prior.

**Appearance dissimilarity loss** As illustrated in Fig. 2, we measure appearance dissimilarity loss between three sequential images. Given inverse depth maps predicted from these images respectively, and two camera extrinsics estimated relative to the second image, we compare the inversely warped first and last images to the second image. We also do it reversely, comparing the inversely warped second image to the other two images. This bidirectional appearance loss allows us to augment the training set with minimum extra computational cost.

Appearance dissimilarity is aggregated through four scales of the output of the Depth-CNN. For the three coarser scales, we use L1 photometric loss; For the last finest scale, we adopt a linear combination of L1 photometric loss and single scale SSIM [30] loss, which is the same as in [13].

**Inverse depth smoothness loss** We use second order gradients as smoothness cost to encourage the depth prediction has flat slopes. To encourage sharpness of the depth prediction, we employ two strategies:
(1) We give different weightings to the smoothness cost at different pixel locations according to the Laplacian of the image intensity. If the Laplacian is higher, meaning the pixel is more likely on an edge or corner, we impose a lower smoothness cost at that location:

$$\mathcal{L}_{\text{prior}}(d_i) = e^{-\nabla^2 \mathcal{I}(\mathbf{x}_i)} \left( |\partial_{xx} d_i| + |\partial_{xy} d_i| + |\partial_{yy} d_i| \right) \tag{16}$$

(2) Instead of directly compute smoothness cost from the final inverse depth estimation, we downsample the inverse depth map by factor 2 to three coarser scales. We aggregate smoothness cost only from the two coarsest scales.

## 5. Experiments

This section extensively evaluates our method of unsupervised training depth estimator from monocular videos. Following the other relevant works [11, 13, 19, 31, 28] we employ KITTI dataset [12] to train our approach. We use the same train/test split used in [7], meaning that we use 28 sequences for training and 28 sequences for testing in the categories of "city", "residential" and "road". Since each of the sequences are captured with a stereo pair, we have 2 monocular sequences giving 56 monocular sequences for training.

**Training set construction** We follow [31] to preprocess the training sequences. To do so, we first remove the static frames by inspecting if the mean magnitude of optical flow is less than 1 pixel. Each training sample is a short clip of 3 sequential video frames resized to $128 \times 416$ pixels.

**Depth-CNN architecture** We use the same network architecture as in [13]. The network resembles a U-Net structure. It consists of an encoder-decoder design with skip connections of intermediate features between the encoder and decoder network. We also output multi-scale inverse depth predictions at the end of the decoder. Since inverse depth is bounded between 0 (at infinity) and the inverse of minimum depth to the camera, we apply sigmoid non-linearity to the network output. Moreover, we multiply the final inverse depth output by 10 to constrain it inside a reasonable range, and add a small value 0.01 to improve numerical stability.

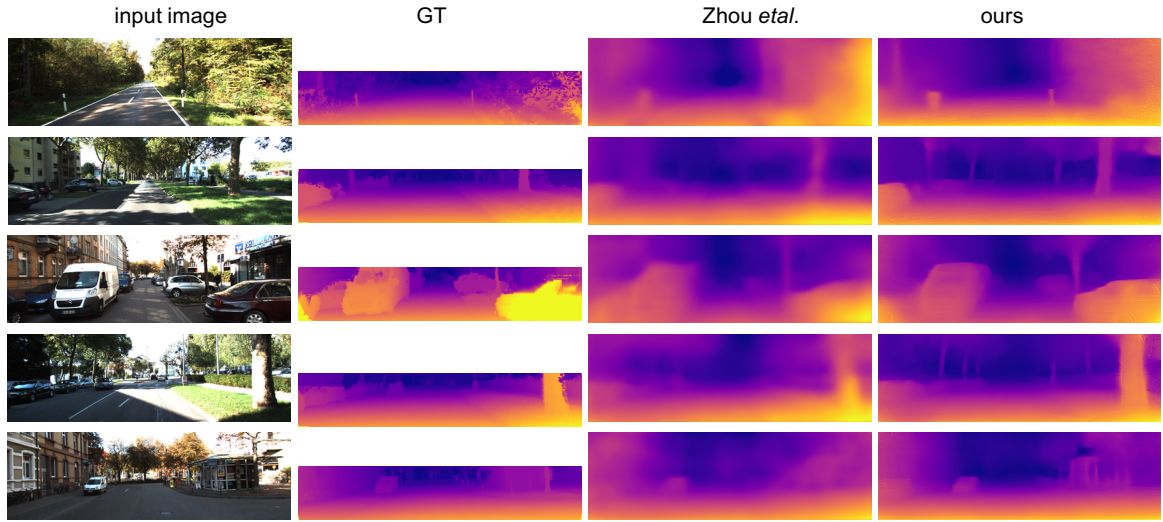

Figure 4. Qualitative results on KITTI test set (Eigen split). Here, groundtruth depth maps are interpolated from sparse point clouds captured by Lidar (2nd column), thus only serves for visualization purpose. Compared to Zhou *et al*. [31] (3rd column), our depth map prediction (last column) preserves more details such as tree trunks, roadway stakes, and gives more precise reconstruction of objects near view, such as the van and tree in the 3rd and 4th rows.

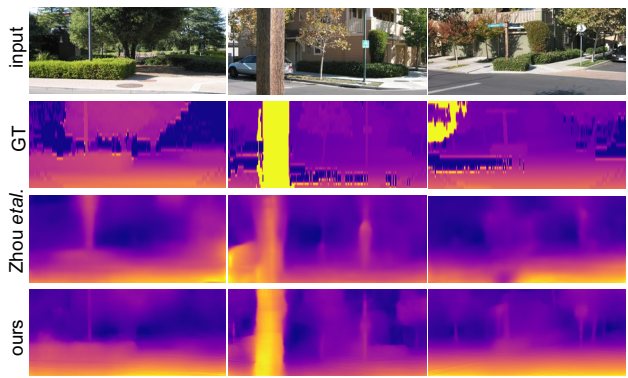

Figure 5. Qualitative results on Make3D dataset. We directly apply our model trained from KITTI without any tuning for this dataset.

**Pose-CNN architecture** We use a similar CNN architecture as in [31], which takes input from a concatenation of three sequential frames. The difference is that instead of outputting rotation as Euler angles, we train our Pose-CNN to predict exponential coordinates.

**Training hyper-parameters** Our training loss is a weighted sum of appearance dissimilarity loss and smoothness loss. We set the weighting for the smoothness loss as 0.01. We empirically found that this weighting parameter offers a desired balance between sharpness and overall structure correctness of the depth prediction.

We train the network through Adam optimizer [16], with a small learning rate of 0.0001, $\beta_1 = 0.9$ and $\beta_2 = 0.999$. Restricted by the non-conventional operation used in our DDVO module, and for the sake of computational efficiency, we set the batch size for training to 1. This compromise on batch size is solely due to implementation issues, not an implication of theory. Empirically, we observe no negative effect on overall performance.

### 5.1. Training configurations

For a complete study of our proposed method, we conduct experiments with four different settings:

**Ours (Baseline)** To understand the effect of our proposed normalization trick in Sec. 2.1, we setup a baseline configuration. We use Pose-CNN as the pose predictor and turn off the inverse depth map normalization. This setup is equivalent to Zhou *et al*. [31]'s method modified with our loss function definition (see Sec. 4) and implementation. Due to the training diverges after 10 epochs (see Fig. 3), we report the result at the 10th epoch.

**Ours (Pose-CNN)** We perform inverse depth normalization, and still use Pose-CNN as the pose predictor. We report result at the 10th epoch.

**Ours (DDVO)**. We replace the Pose-CNN with our proposed DDVO module. The DDVO is initialized with the identity pose. To account for large motion in the training set, we set the module to run through five scales in a coarse-to-fine manner. We train the model from scratch, and result is reported at the 10th epoch.

**Ours (Pose-CNN + DDVO)**. We use the Depth-CNN and Pose-CNN from "ours (Pose-CNN)" as pretrained model. Then we fix the Pose-CNN, use its output as pose initialization for our DDVO module, and fine-tune the Depth-CNN for 2 epochs. Since Pose-CNN already gives us a roughly acceptable estimation, DDVO does not need to account for large motion in this case. Hence, to speed up training, instead of performing coarse-to-fine, we directly run DDVO over the finest scale.

### 5.2. Results on KITTI

In this section, we evaluate performance on the same 697 KITTI images from the Eigen's test split [7]. We compute all the errors from the same cropped region using the same performance measures as in [11, 31, 13] and without capping to a maximum depth. Our method cannot recover the true scale of the scene due to the monocular settings of the problem. Therefore, we align the scale of our prediction to the groundtruth through multiplying the former by $\tilde{s} = \text{median}(D_{\text{gt}})/\text{median}(D_{\text{predict}})$ as in [31].

**Effect of depth map normalization**. We compare the result of "ours (Baseline)" to "ours (Pose-CNN)" since these two configurations only differ in whether inverse depth normalization is performed or not (see section 2.1). As shown in Table 1, the simple normalization approach gets a significant improvement in all the measures using the same architecture. Moreover, with the results of depth normalization, "Ours (Pose-CNN)" not only beats Zhou *et al.* [31]'s work by a large margin, and also gets very close results to the one trained on rectified stereo pairs by Godard *et al.* [13].

**DDVO vs Pose-CNN** We also test the effect of changing the Pose-CNN module by the DDVO module proposed in section 3. In this case, we have no parameter to learn in the pose estimation module. In Table 1 we show the results with label "ours (DDVO)". As the DDVO module performs Gauss-Newton optimization, we initialize the pose between each pair of training images as the identity. On the other hand, Gauss-Newton approaches are sensible to the initialization point. This is the reason for getting slightly worse results with DDVO pose estimation ("ours (DDVO)") than with a learned Pose-CNN pose estimator ("ours (Pose-CNN)"). In any case the results are still very close to the stereo pairs training method of Godard *et al.* [13].

Finally, in order to improve the pose initialization point in the DDVO module we have experimented a new way of training. The pose between a pair of images in a video sequence is initialized by a pretrained Pose-CNN module and then refined with the DDVO module. In Table 1 we show that this strategy ("ours (Pose-CNN+DDVO)") improves both "ours (Pose-CNN)" and "ours (DDVO)", and gives the best results of the unsupervised trained methods trained on KITTI dataset. Qualitative results from this ap-proach are visualized in Fig. 4.

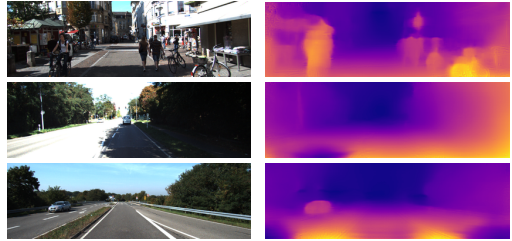

Figure 6. Failure cases of our method. Our method is sensitive to 1) dynamic scenes like pedestrians and cyclists (1st row); 2) overexposure of the road, which cause large texture-less regions (2nd row); 3) vast open areas with large texture-less regions (3rd row).

**Pretrain on Cityscapes** As shown in [31, 13] pretraining the model on a larger dataset, *e.g.* Cityscapes [5], and then finetuning on KITTI improve the results. Hence, we also perform a similar experiment. First, we train on Cityscapes with "ours (Pose-CNN)" architecture for 10 epochs, then finetune the model on KITTI with "ours (Pose-CNN)" for the first 5 epochs and "ours (Pose-CNN + DDVO)" for the next 3 epochs. We observe a slight performance improvement over training only on KITTI but still outperformed by Godard *et al.* [13]. This states a bottleneck of our approach, where our method is not modeling the non-rigid part of the scene. This prevents us to get more benefit from a dataset with a lot of objects in motion.

### 5.3. Results on Make3D

We directly apply our model trained from KITTI (Ours (Pose-CNN + DDVO)) to the Make3D test set [26]. Due to the difference in camera aspect ratio, we only evaluate our method on the central crop of Make3D's test images. We find that our method generalizes moderately well on this dataset. As shown in Fig. 5, our method can recover details such as the road sign in distance, and wooden pole in close view. Table 2 shows that our method achieves similar or better results compared to other state of the art approaches.

### 6. Discussion

We have found that, as in monocular SLAM algorithms [10], scale ambiguity has to be taken into account when training with monocular videos. As shown in Fig. 3 this was a missing point in previous approaches training from monocular videos. Additionally, we have shown that Direct Visual Odometry can be used to estimate the relative camera pose between frames instead of learning a predictor with a CNN (Pose-CNN). This strategy needs less parameters to learn and potentially less images to train. We plan to investigate the number of images needed in the future. Finally, we have found that DDVO and Pose-CNN pose prediction modules can be improved with a hybrid architecture

| Method | training set | abs rel | sq rel | RMSE | RMSE(log) | $\delta < 1.25$ | $\delta < 1.25^2$ | $\delta < 1.25^3$ |
|---|---|---|---|---|---|---|---|---|
| Eigen et al. [7] | K (D) | 0.203 | 1.548 | 6.307 | 0.282 | 0.702 | 0.890 | 0.958 |
| Liu et al. [23] | K (D) | 0.202 | 1.614 | 6.523 | 0.275 | 0.678 | 0.895 | 0.965 |
| Godard et al. [13] | K (B) | **0.148** | 1.344 | 5.927 | 0.247 | 0.803 | 0.922 | 0.964 |
| Zhou et al. [31] | K (M) | 0.208 | 1.768 | 6.856 | 0.283 | 0.678 | 0.885 | 0.957 |
| ours (baseline) | K (M) | 0.213 | 3.770 | 6.925 | 0.294 | 0.758 | 0.909 | 0.958 |
| ours (Pose-CNN) | K (M) | 0.155 | 1.193 | 5.613 | 0.229 | 0.797 | 0.935 | 0.975 |
| ours (DDVO) | K (M) | 0.159 | 1.347 | 5.789 | 0.234 | 0.796 | 0.933 | 0.973 |
| **ours** (Pose-CNN+DDVO) | K (M) | 0.151 | 1.257 | 5.583 | 0.228 | 0.810 | 0.936 | 0.974 |
| Godard [13] | CS+K (B) | 0.124 | 1.076 | 5.311 | 0.219 | 0.847 | 0.942 | 0.973 |
| Zhou et al. [31] | CS+K (M) | 0.198 | 1.836 | 6.565 | 0.275 | 0.718 | 0.901 | 0.960 |
| **ours** (Pose-CNN+DDVO) | CS+K (M) | 0.148 | 1.187 | 5.496 | 0.226 | 0.812 | 0.938 | 0.975 |

Table 1. Evaluation of depth map accuracy on KITTI test set. The methods trained over KITTI dataset are denoted by K, and also pretrained on Cityscapes are denoted by CS+K. We use (D) to represent methods trained with depth supervision, (B) to refer methods trained using rectified binocular image pairs and (M) to denote methods trained on monocular video sequences. We show in **bold** the overall best results trained only on KITTI, and with a blue box to highlight the best results within the (M) methods. Gordard and Zhou's results are taken directly from [31].

| Method | Supervision | Abs Rel | Sq Rel | RMSE | RMSE(log) |
|---|---|---|---|---|---|
| Train set mean | depth | 0.876 | 13.98 | 12.27 | 0.307 |
| Karsch et al. [14] | depth | 0.428 | 5.079 | 8.389 | 0.149 |
| Liu et al. [24] | depth | 0.475 | 6.562 | 10.05 | 0.165 |
| Laina et al. [20] | depth | 0.204 | 1.840 | 5.683 | 0.084 |
| Godard et al. [13] | pose | 0.544 | 10.94 | 11.76 | 0.193 |
| Zhou et al. [31] | **none** | 0.383 | 5.321 | 10.47 | 0.478 |
| **ours** | **none** | 0.387 | 4.720 | 8.09 | 0.204 |

Table 2. Results on the Make3D dataset [26]. Following Zhou et al. [31], we do not use any of the Make3D data for training. We directly apply the model trained on KITTI to the test set. Following the evaluation protocol of [31], we first match the scale of our depth prediction to the groundtruth, then the errors are computed only for pixels in a central image crop with groundtruth depth less than 70 meters.

of Pose-CNN initialization and DDVO pose refinement giving the best results.

The current major bottleneck for our approach is that we're not modeling the non-rigidity of the world. As shown in Fig. 6, our current method does not perform well for articulated objects like bikers and pedestrians. Possible future work for this is to incorporate techniques from non-rigid SfM e.g. [18] to the pipeline.

# Learning Depth from Monocular Videos using Direct Methods
– Supplementary Material

## 1. Differentiable vs non-differentiabel direct visual odometry module

As an alternative approach, instead of end-to-end optimizing the objective defined in Eq. 6, we could employ an EM style optimization procedure – first estimate the camera pose $\mathbf{p}^*$ given current depth prediction; then with the pose estimation fixed, we minimize the loss with respect to depth estimation. This is summarized as iteratively performing the following two steps:

$$\mathbf{p}^* = f_p(f_d(\mathcal{I};\theta_d), \mathcal{I}_1, ..., \mathcal{I}_n). \qquad (1)$$

$$\min_{\theta_d} \mathcal{L}_{\text{ap.}}\left(f_d(\mathcal{I};\theta_d), \mathbf{p}^*\right) + \mathcal{L}_{\text{prior}}\left(f_d(\mathcal{I};\theta_d)\right). \qquad (2)$$

The implication of this approach is that we no longer need to do back propagation through the direct visual odometry module, since in the second step we pretend $\mathbf{p}^*$ to be some constant. Hereby, we refer this approach as using *non-differentiable* direct visual odometry module (denoted as "DVO" in Fig. 1).

To test this approach, we did a toy experiment which trains the depth estimator to overfit a short video clip. As shown in Fig. 1, this approach (referred as "DVO") converges slower (see the blue curve in right) than the *differentiable* direct visual odometry module("DDVO", red curve) proposed in the paper, and it stucks in a poor local minimum(visualized in bottom left).

Therefore, we conclude that using *differentiable* DVO module is necessary for training a depth estimator from monocular videos.

## 2. Estimate pose from depth prediction

Tateno *et al*. [1] demonstrates that one plausible usage of monocular depth estimator is to initialize the depth estimation for the visual SLAM systems. Hence, to test the implication of the improvement in depth map accuracy under this usage scenario, we run the direct visual odmetry algorithm(described in section 3.1) over depth maps predicted by Zhou *et al*. [2] and our method. Through inspecting the pose errors (measured by absolute trajectory error), we find that the improvement in our depth map accuracy results in significantly better pose estimation (see Table 2).

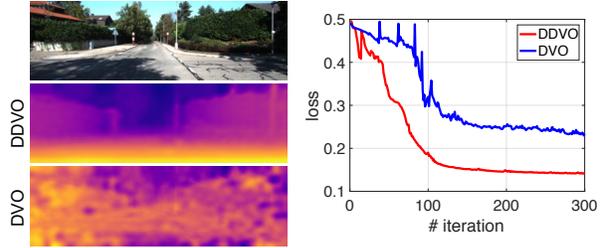

Figure 1. Compare training with *differentiable* DVO ("DDVO") to *non-differentiable* DVO module ("DVO") on one short clip of video. right: "DDVO"(red) converges faster and reaches better local minimum compared to "DVO"(blue); left-mid: "DDVO" gives a good reconstruction of the scene; left-bottom: "DVO" results in wrong depth estimation.

| Method | Seq. 09 | Seq. 10 |
|---|---|---|
| ORB-SLAM(full) | 0.014±0.008 | 0.012±0.011 |
| ORB-SLAM(short) | 0.064±0.141 | 0.064±0.130 |
| Zhou *et al*. [2] | 0.063±0.126 | 0.085±0.115 |
| **Ours**(Pose-CNN + DDVO) | **0.045**±**0.108** | **0.033**±**0.074** |

Table 1. Absolute Trajectory Error (ATE) on the KITTI odometry split averaged over all 5-frame snippets. ORB-SLAM results are copied from [2]. "full" means ORB-SLAM is performed on the full video while "short" means it's only performed on the 5-frame snippets. For both Zhou *et al*. and ours, we use the Depth-CNNs trained from the combined dataset of Cityscapes and KITI.

## 3. Video demo on KITTI sequence

We attached a video demo comparing our depth estimation to Zhou *et al*. [2] over a sequence from KITTI Eigen test split. Both methods are pre-trained on Cityscapes and finetuned on KITTI.